\documentclass{article}

\usepackage{arxiv}

\usepackage{microtype}
\usepackage{tikz}
\usepackage{amsmath}
\usepackage{algorithm} 
\usepackage{algpseudocode} 
\usepackage{graphicx}
\usepackage{booktabs} 
\usepackage{subcaption}
\usepackage{float}
\usepackage{natbib}
\usetikzlibrary{positioning, shapes.geometric}

\DeclareMathOperator*{\argmin}{arg\,min}

\usepackage{hyperref}

\usepackage{amsmath}
\usepackage{amssymb}
\usepackage{mathtools}
\usepackage{amsthm}

\usepackage[capitalize,noabbrev]{cleveref}

\theoremstyle{plain}

\theoremstyle{definition}

\theoremstyle{remark}

\usepackage[textsize=tiny]{todonotes}

\title{Adversarial Debiasing for Unbiased Parameter Recovery}

\author{
  \href{https://orcid.org/0000-0002-2641-0636}{\includegraphics[scale=0.06]{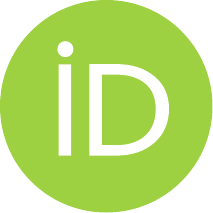}\hspace{1mm}Luke C Sanford}\thanks{Sanford, Ayers, and Gordon share co-first authorship of this work. Sanford and Gordon conceived of the idea, Ayers and Gordon led development of the debiasing approach and software implementation, Ayers and Gordon developed the statistical proofs, Ayers and Stone led the data collection, Sanford led the bias test and simulation, Sanford and Ayers led and all authors contributed to writing the manuscript.}\\
  School of the Environment\\
  Yale University\\
  New Haven, CT \\
  \texttt{luke.sanford@yale.edu} \\
  \And
  \href{https://orcid.org/0009-0005-5762-8266}{\includegraphics[scale=0.06]{orcid.pdf}\hspace{1mm}Megan Ayers}\footnotemark[1]\\
  Department of Statistics and Data Science\\
  Yale University\\
  New Haven, CT \\
  \And
  \href{https://orcid.org/0000-0001-8241-3754}{\includegraphics[scale=0.06]{orcid.pdf}\hspace{1mm}Matthew Gordon}\footnotemark[1]\\
  Paris School of Economics\\
  Paris, France \\
  \And
  \href{https://orcid.org/0000-0003-4275-1294}{\includegraphics[scale=0.06]{orcid.pdf}\hspace{1mm}Eliana Stone}\\
  School of the Environment\\
  Yale University\\
  New Haven, CT \\
}

\renewcommand{\shorttitle}{Adversarial Debiasing}

\hypersetup{
pdftitle={Adversarial Debiasing},
pdfsubject={ArXiV preprint},
pdfauthor={Luke C Sanford, Megan Ayers, Matthew Gordon, Eliana Stone},
pdfkeywords={Machine Learning, Remote Sensing, Adversarial Debiasing},
}

\begin{document}
\maketitle

\begin{abstract}
	Advances in machine learning and the increasing availability of high-dimensional data have led to the proliferation of social science research that uses the predictions of machine learning models as proxies for measures of human activity or environmental outcomes. However, prediction errors from machine learning models can lead to bias in the estimates of regression coefficients. In this paper, we show how this bias can arise, propose a test for detecting bias, and demonstrate the use of an adversarial machine learning algorithm in order to de-bias predictions. These methods are applicable to any setting where machine-learned predictions are the dependent variable in a regression. We conduct simulations and empirical exercises using ground truth and satellite data on forest cover in Africa. Using the predictions from a naive machine learning model leads to biased parameter estimates, while the predictions from the adversarial model recover the true coefficients. 
\end{abstract}


\section{Introduction}
\label{introduction}

Efforts to combat climate change, alleviate poverty, and improve food security rely on studies that evaluate the effectiveness of policy interventions. These studies increasingly rely on satellite remote sensing data and machine learning methods to measure outcomes (biodiversity, forest carbon, wealth, agricultural productivity) of interest. Converting multi-spectral, multi-temporal, and spatial data into these dependent variables previously relied on simple heuristics for dimension reduction (for example, determining the date of max greenness). More recently, researchers have turned to machine learning methods paired with labeled data to generate more precise measures of these outcomes. With the increasing availability of remote sensing data, decreasing cost of computation, and continuing improvement of machine learning tools, we are in the midst of a measurement revolution in social, economic, and environmental sciences.

These advances in measurement derive from the field of remote sensing, which generally develops tools to improve measurement of particular outcomes, $Y$ using satellite data. These approaches generate $\widehat{Y}_i(k_i) = Y_i + \nu_i$, where $k$ are the satellite imagery features we use to train our machine learning model and $\nu_i$ is the measurement error for a given observation. This model, $f(k_i, w)$, finds weights $w$ that minimize some function of $\nu$ in the labeled data, for example the sum of squared error: $\sum_j\nu_j^2 = \sum_j(\widehat{Y}_j(k_j) - Y_j)^2$. While this is a sensible approach for descriptive tasks (how much deforestation occurred in the Amazon in 2024?), it can generate bias when $\widehat{Y}$ is used as a dependent variable for causal inference tasks (how much deforestation in the Amazon was caused by the expansion of road networks?).

This bias can occur because of differential measurement error across some independent variable, for example, if forest cover is over-estimated far from roads and under-estimated near roads. Then, researchers who estimate the effect  $\beta$ of roads $X_i$ on forest cover $\widehat{Y_i}$ according to the model  $Y_i = \alpha + \beta X_i + e_i$ will obtain in expectation:
\begin{equation} \label{eq:bias}
    \mathbb{E}[\widehat{\beta}] = \beta + \frac{Cov(e, X)}{Var(X)} + \frac{Cov(\nu, X)}{Var(X)}.
\end{equation}
$\beta$ is the true effect, the second term is standard omitted variable bias, and the third term is measurement error bias caused by using predicted forest cover as the dependent variable rather than the true values. This is problematic when there is a correlation between the independent variable and the measurement error. We describe why we expect this type of correlation to be common in studies that use remote sensing data to generate dependent variables. We then construct a statistical test for bias arising from measurement error bias and develop a power calculation that allows researchers to estimate the number of ground truth points they will need to acquire.

While other work in this area takes the model predictions $\widehat{Y}$ as given and seeks to correct them using ground truth labels in the form of $\tilde{Y} = m(\widehat{Y}, Y, X, k)$, we develop a class of models that generate $\widehat{Y}$ such that $Cov(\nu, X) = 0$. To achieve this, we borrow an approach from the algorithmic fairness literature where prediction errors are constrained to be uncorrelated with legally protected categories like race or gender \cite{zhang_mitigating_2018}. Rather than having the adversarial component of the model attempt to predict one or more protected categories, we task it with predicting treatment status (i.e. the independent variable of interest $X$). This approach generates predictions that are both precise and unbiased for the researcher's causal task.

\begin{figure}[ht]
\vskip 0.2in
\begin{center}
\centerline{\includegraphics[width=\linewidth]{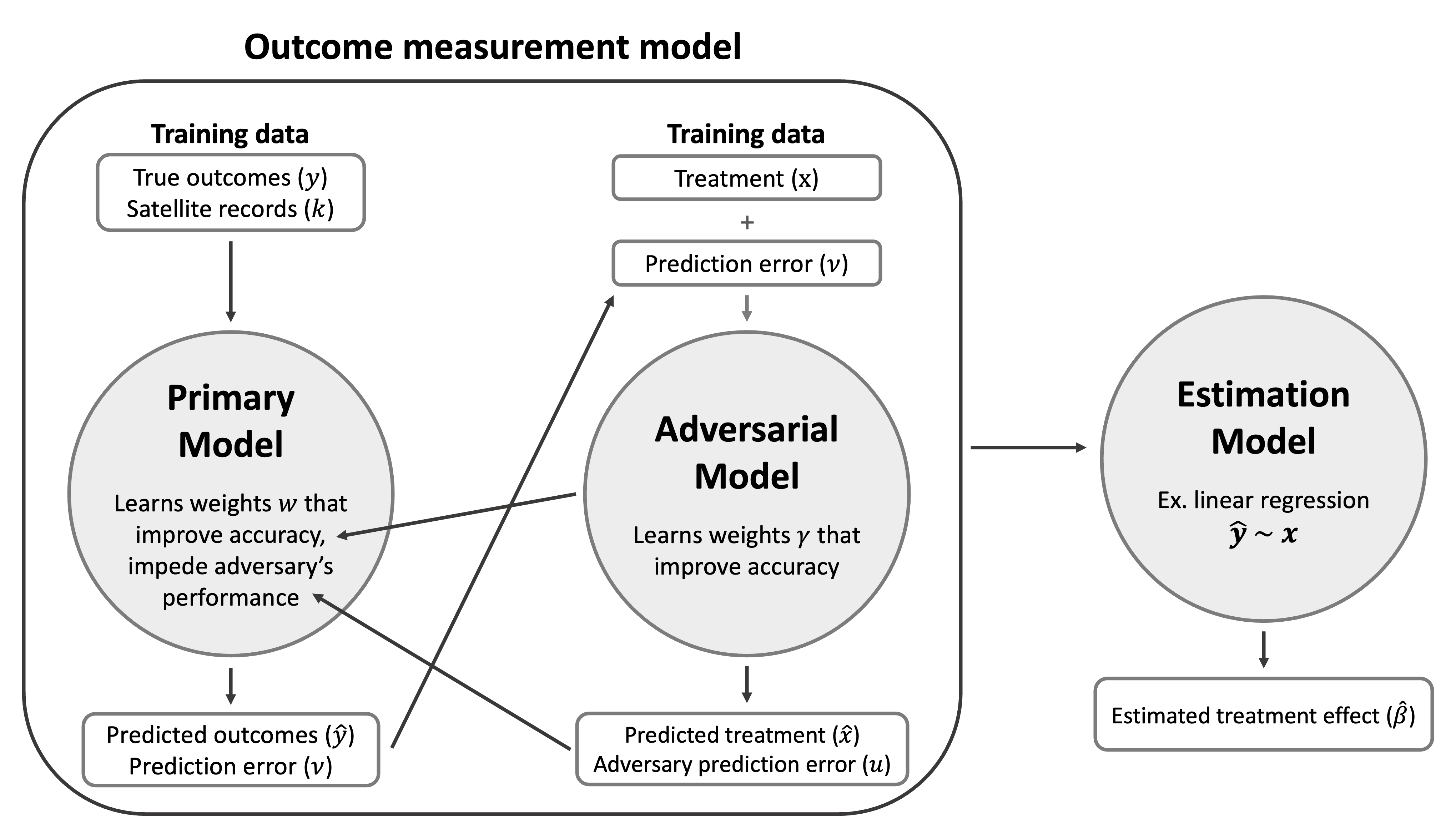}}
\caption{The model architecture of the debiasing approach}
\label{graphical_algo}
\end{center}
\vskip -0.2in
\end{figure}

\subsection{Contributions}

The main contribution of the paper is a novel approach to the increasingly recognized issue of using machine-learned proxies for causal tasks. While all other approaches we know of attempt to correct off-the-shelf measurements or measurements from a fixed model, we propose an adversarial component that can be added to any measurement model to generate predictions that avoid measurement error bias. This eliminates extra bias correction steps in the causal estimation process, which can decrease precision. This natively fits workflows in which researchers are constructing their own measurement models--something we expect to be more common in remote sensing with the advent of geospatial foundation models. We also develop what to our knowledge is the first statistical test for measurement error bias and the first power calculation workflow that guides researchers in understanding how much ground truth data they will need to label to ensure their results are robust to measurement error bias. 

\subsection{Related Work} \label{litreview}
This paper contributes to a growing literature that has begun to document the problem of non-random measurement error in machine-learning models (see \cite{jain_benefits_2020} for a review\footnote{For topic specific reviews, see \cite{balboni_economics_2022} on deforestation, \cite{gibson_which_2021} and \cite{bluhm_what_2022} on night lights, and \cite{fowlie_bringing_2019} on air pollution.}). A number of recent papers propose econometric estimators that can correct for the non-classical measurement error in some cases. \citet{alix-garcia_remotely_2022} proposes a misclassification model that requires users to specify the variables that may induce measurement error (e.g. cloud cover, satellite angle). \citet{proctor_parameter_2023} suggest a multiple imputation approach that may be sensitive to functional form specifications. \citet{egami_using_2023} address an analagous problem in the NLP realm, developing a method similar to \cite{proctor_parameter_2023} for using gold-standard labeled data to adjust labels provided by an LLM. \citet{zhang_how_2021}
propose analytical bias correction techniques and MLE estimators to handle attenuation bias when classifier predictions are used as outcomes, but assume non-differential measurement error with respect to regressors. \citet{torchiana_improving_2023} provide an approach based on hidden Markov models to correct for misclassification bias that does not require ground-truth data but requires a stronger set of structural assumptions about data generating processes. In contrast, bias estimation and correction methods \citep[such as prediction-powered inference introduced in][]{angelopoulos_prediction-powered_2023} and our proposed adversarial debiasing method make i.i.d sampling assumptions but do not require assumptions of functional forms or specific sources of measurement error, and are designed to target differential measurement error.



\section{Adversarial Debiasing}
\label{method}

\subsection{Detecting Bias}
Unlike for omitted variable bias, an estimate of measurement error bias is directly obtainable if researchers have access to or can generate some ground-truth values of $Y$ (for example, by visually interpreting high-resolution satellite imagery). 
Let $j \in J$ index observations in the labeled set. Consider the regression:

\begin{equation} \label{eq:biastest}
    \nu_j = X_j \gamma + u_j,
\end{equation}

with $X$ as a vector of independent variables, and $\nu_j = \widehat{Y} - Y$ is the prediction error. Our estimate of $\gamma$ will be $\widehat{\gamma} = (X'X)^{-1}X'\nu$, which will converge to the multivariate analog of the bias term in equation \ref{eq:bias} under the assumption that the labeled set is representative of the evaluation set. This shows that a very simple regression coefficient can be used to estimate downstream bias upon deployment of the model over the entire evaluation set.


Estimates of the standard errors of $\widehat{\gamma}$ can be used to test whether the bias is significantly different from zero, or to rule out biases greater than a certain size, though standard errors likely need to be adjusted for spatial or serial correlation, and/or bootstrapped to account for all uncertainty associated with model-derived outcomes. It is simple to adapt standard power calculations to estimate a `minimum detectable bias' given a certain number of observations and an estimate of the standard error of $\widehat{\gamma}$. Researchers can then estimate the number of labeled observations which will likely be necessary to rule out some amount of measurement error bias. We demonstrate this procedure in our first two applications.

Direct estimates of bias can also be used to perform a `bias correction' on estimates of $\widehat{\beta}$ from equation \ref{eq:bias}:

\begin{equation} \label{bias_corr}
    \widehat{\beta}_c = \widehat{\beta} - \widehat{\gamma}.
\end{equation}

This is a special case of the prediction powered inference approach introduced in \citet{angelopoulos_prediction-powered_2023}. In expectation, \ref{bias_corr} will coincide with the true value of $\beta$ when there is no endogeneity bias and the representative labeled sample assumption holds. An estimate of the standard error of $\widehat{\beta}_{c}$ can be produced by a normal approximation as described in \citet{angelopoulos_prediction-powered_2023} or with a bootstrap procedure. 

There are two main benefits to this approach. First, $\widehat{\beta}$ and $\widehat{\gamma}$ can be generated with off-the-shelf predictions and a relatively small amount of ground-truth data, which is straightforward for those who do not want to generate a custom machine learning model for their research question as we describe in the following section. Second, we expect this approach to outperform the following approach in settings constrained to a very small number of ground truth observations when it will be difficult to train an accurate machine learning model with many parameters. There are shortcomings of this estimator, however. The estimation uncertainty of $\widehat{\gamma}$ inflates the variance of $\widehat{\beta}_{c}$. 
Secondly, this approach takes a set of $\widehat{Y}$ predictions as given, which fixes the precision of $\widehat{\beta}$.
In the next section, we describe how to obtain precise predictions of $\widehat{Y}$ without differential measurement error for a given estimation problem.

\subsection{Adversarial Model}



Rather than performing post-hoc bias corrections, we investigate outcome prediction models that directly avoid differential measurement error by construction. Given the form of bias described in the previous section, we investigate adding a penalty to standard loss functions to disincentive prediction errors that are correlated with $X$. We seek a model with parameters $w^*$ that satisfy:

\begin{equation}
\begin{gathered} \label{constrainedloss}
    \omega^* = \argmin_\omega L_p(\widehat{Y}(\omega), Y, k)\\
    \text{such that } Cov(X, Y - \widehat{Y}(\omega)) \approx 0, 
\end{gathered}    
\end{equation}

where $L_p$ is a standard loss function, such as mean squared error. Intuitively, the constraint on the loss function is a requirement that the measurement errors contain as little `information' as possible about $X$. When linear regression is used as the downstream estimation model, we are specifically concerned with `information' in the form of a linear relationship. Generalizing across estimation strategy, we can consider a `no information' constraint as satisfied when a second machine learning model is not able to predict $X$ accurately from the errors. Call such a model an adversary and define its loss function as $L_a(\hat{X}(\gamma), X, Y, \widehat{Y}(\omega))$, with model weights $\gamma$.\footnote{$\gamma$ is intentionally chosen to be the same variable as the coefficient in the bias test above for reasons that are about to become clear.} This approach is called adversarial debiasing, and was first proposed by \citet{zhang_mitigating_2018} to debias machine learning model predictions with respect to race or gender.

Adversarial debiasing models are trained to minimize $L_p$, while maximizing $L_a$, subject to the adversary choosing $\gamma$ in such a way as to minimize $L_a$ (Figure \ref{graphical_algo}). Formally, this can be written as the following objective function for training the primary prediction model $P$:

\begin{equation}
    \begin{gathered} \label{adversaryloss}
    \min_\omega L_p(\widehat{Y}(\omega), Y, k) - \alpha L_a(X, Y, \widehat{Y}(\omega), \gamma, k)
\end{gathered}
\end{equation}

where $\gamma \in \text{argmin } L_a(X, Y, \widehat{Y}(\omega), \gamma, k)$ and $\alpha$ controls the weight on the adversary's loss function, which must be chosen by the researcher (e.g. by cross fitting). Now consider if the adversary model is a linear model of the exact form of the bias test above:

\begin{gather} \label{linearadversary}
    \nu_j = X_j \gamma + \epsilon_j
\end{gather}
with loss function as the sum of squared errors. The loss function of this adversary is minimized with respect to $\gamma$ when $\gamma = (X'X)^{-1}X'\nu$. The primary model will try to choose $\omega$ such that the prediction errors $\nu$ maximize the adversary's loss function:
\begin{gather}
    L_a = \frac{1}{N}\|\nu - X(X'X)^{-1}X'\nu\|^2, \nonumber
\end{gather}
while balancing this task with overall accuracy.

Consider prediction errors at step $t$ and $t + 1$ of training: $\nu_t$ and $\nu_{t + 1}$. Holding accuracy of the primary model predictions constant, moving in the opposite direction of the adversarial loss gradient means the bias at time $t+1$ will be smaller than at time $t$, $|\gamma_{t + 1}| < |\gamma_t|$, assuming that treatment $X$ is univariate (Proof in Appendix \ref{sec:appendixa}).
Thus, if $\alpha$ is sufficiently high, choosing $\omega$ to optimize equation \ref{adversaryloss} implies that the bias term will shrink if accuracy is stable. In practice, the model will balance the goal of maximizing the adversary's loss with the goal of overall prediction accuracy. 

Another simple approach in the special case of a simple linear regression estimation model is to penalize the covariance of $X$ and $\nu$ directly in the loss function:
\begin{equation}
    \min_\omega L_p(\widehat{Y}(\omega), Y, k) - \alpha \Bigl| \sum_j (x_j - \bar{x}) \nu_j \Bigr|.
\end{equation}
This method is also applicable to bias terms that can be derived for regressions with controls, or an instrumental variables setting, with slight modifications (details in Appendix \ref{sec:appendixb}). 

In our applications, we experiment with both approaches. For both methods, the choice of $\alpha$ is important. With too low of a value of $\alpha$ the model does not effectively debias the results and minimizes squared prediction error instead of maximizing the adversarial loss. However, when $\alpha$ is too high the model may produce inaccurate but idiosyncratic measurements to inflate the adversary's loss (a coin flip to determine $\widehat{Y}$ would satisfy our bias constraint, but the predictions would be useless). We use cross fitting within the labeled data to set $\alpha$ so that overall prediction error and bias can be examined for different values of $\alpha$.

\subsection{Standard Errors} \label{sec:se}

When using predicted outcomes as the dependent variable in a linear regression, the variance of the estimated coefficients can be expressed as:

\begin{align}
    Var(\widehat{\beta}) =& Var\left( (X'X)^{-1} X'\widehat{Y} \right)\\  \nonumber
    =& Var\left( (X'X)^{-1} X'e \right) + \\ \nonumber
    & Var\left( (X'X)^{-1} X'\nu \right) + \\ \nonumber
    & 2 Cov\left( (X'X)^{-1} X'e, (X'X)^{-1} X'\nu \right) \nonumber
\end{align}

Holding the covariance term constant (we do not focus on traditional confounding error in this work, so consider this term to be 0), larger prediction errors will inflate the variance of estimates of $\beta$ through the second term, compared to an ideal case where all data is labeled. This highlights the potential efficiency gains from adversarial debiasing compared to bias correction. The bias correction method takes the set of prediction errors as given, and these are possibly from a non-optimized model. Formally, if adversarial model prediction errors are $\nu_a$, standard model prediction errors are $\nu$, and we assume covariance terms are 0, we have that

\begin{align}  
    \mathrm{Var}(\widehat{\beta}) =&  \mathrm{Var}((X'X)^{-1}X'\nu_a) + \mathrm{Var}((X'X)^{-1}X'e) \nonumber \\
    \mathrm{Var}(\widehat{\beta}_c) =& \mathrm{Var}((X'X)^{-1}X'\nu) + \mathrm{Var}((X'X)^{-1}X'e) \nonumber \\ 
    & + \mathrm{Var}((\tilde{X}'\tilde{X})^{-1}\tilde{X}'u),
\end{align}

where here $\tilde{X}$ and $X$ distinguish data from the labeled data set the evaluation set respectively. Depending on the relative precision of the debiasing model predictions and the predictions used for the bias correction approach, and the number of labeled samples, the debiasing approach can result in efficiency gains. Assuming that debiased model predictions have a similar variance to the off-the-shelf predictions, the bias correction approach will be less efficient due to the additional uncertainty in estimating the bias term.


In practice we can use the typical OLS standard errors or heteroskedasticity-consistent standard errors to estimate the variance of $\widehat{\beta}$ under-sampling uncertainty after performing adversarial debiasing. However, this doesn't take into account uncertainty about the prediction model itself. If the labeled training set was sampled again by the same mechanism and from the same source, it might lead to a different trained model, a different set of outcome measurements, and a different estimate of $\widehat{\beta}$. This is particularly relevant for models that often build randomization into the training process, such as neural networks or random forests. Even fixing a training sample, re-training these models with different seeds results in different estimates. Most studies using machine-learning measurements as outcome variables do not account for model-based uncertainty in their reported standard errors. One approach to account for this is to bootstrap over the training of the machine learning model. This approach is computationally intensive, since it requires many trainings of a potentially complex machine learning model. However, our simulations show that it can be important for capturing uncertainty in practice.


\section{Applications}
\label{applications}
\begin{figure}[t] 
\centering
\resizebox{\columnwidth}{!}{%
    \begin{tikzpicture}[remember picture]
        \node[anchor=south west,inner sep=0] (leftplot) at (0,0)
        {\includegraphics[width=1.0\textwidth]{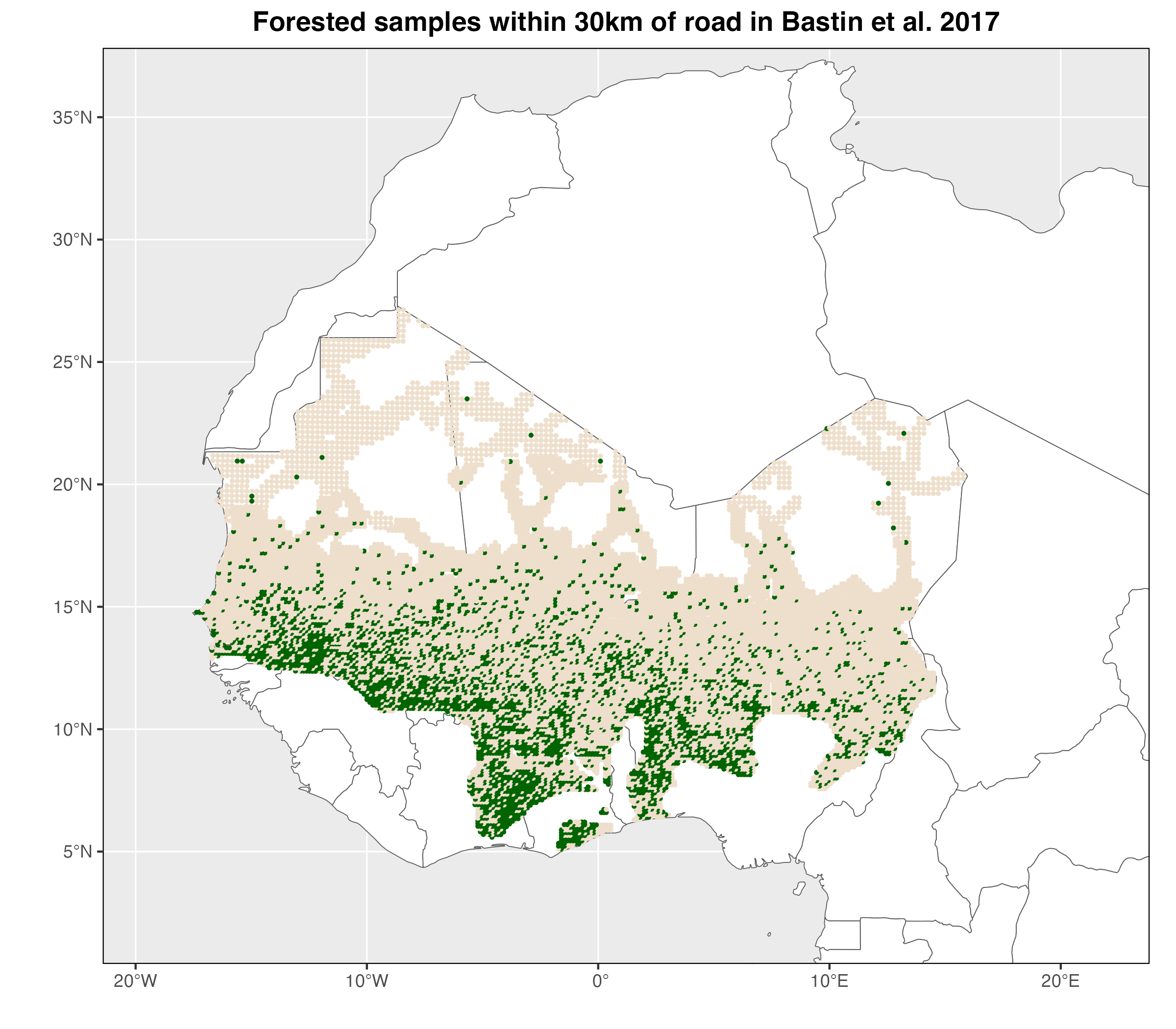}};

        \node[anchor=south west,rectangle,draw=black,minimum width=2cm,%
              minimum height=1.5cm,fill=white,inner sep=0] (viewport) 
              at (leftplot.south west)
        {\includegraphics[width=0.25\textwidth]{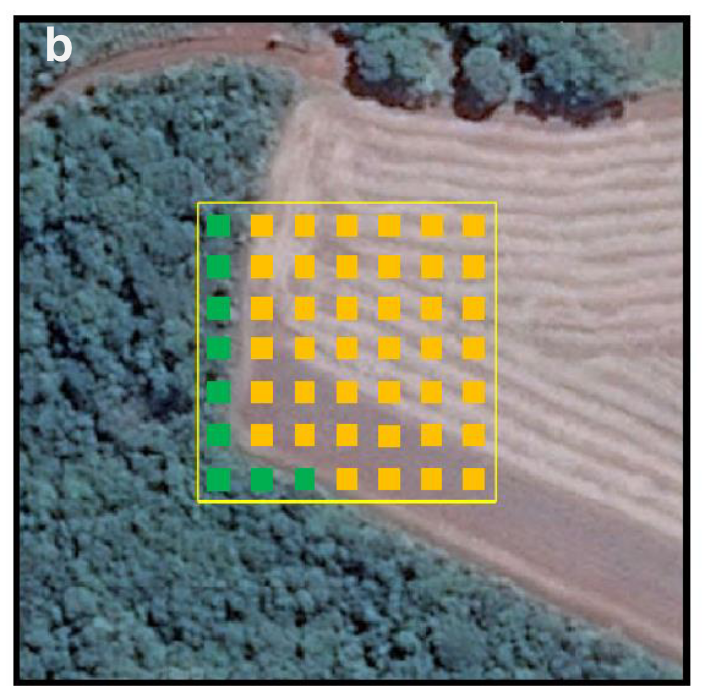}};

        \node[below=2mm of viewport,font=\scriptsize]
        {\tiny{Bastin et al. (2017) Fig S16b.}};
    \end{tikzpicture}%
}
\caption{Colored areas show labeled pixels from \citet{bastin_extent_2017} --- 
green for forested and beige for non-forested. Inset shows 
an example of how percent forested labels were generated 
from high resolution satellite data.}
\label{figure:labeling}
\end{figure}

To demonstrate the efficacy of our approach, we conduct simulations and empirical exercises predicting forest cover in settings where we also have access to ground truth data. We use a dataset of 20,621 points in West Africa (Bastin et al. 2017); each point is labeled with percent treecover and a binary forest/not forest label.  The data was originally collected in part to show the biases of the widely used \citet{hansen_high-resolution_2013} data set in dryland areas. Researchers used high-resolution imagery from between 2011 and 2015 to label the data --- shown in Figure \ref{figure:labeling}. We use a sample of the data in West Africa within 30 km of a road (see Figure \ref{figure:labeling}).

As inputs to our machine learning models, we use data from the Landsat 7 ETM sensor. This sensor records the surface reflectance of light at visible, near-infrared, and infrared wavelengths (called `bands' in the remote sensing literature) at a 30-meter resolution. We generate three popular indices from these bands: Normalized Differenced Vegetation Index, Normalized Differenced Built Index, and Enhanced Vegetation Index. Over the course of a year, each location is observed up to 28 times (cloudiness obscures locations in some areas at some times). We take the 25th, 50th and 75th percentiles of each of the eight bands and three indices, and use those 24 variables as inputs. This feature-engineering strategy mirrors the approach in \citet{hansen_high-resolution_2013}. With a simple 1-layer neural network (a logistic regression) we are able to predict binary forest cover using this data with 75\% accuracy.

\subsection{Simulation} \label{sec:sim}

All of the following empirical exercises take the following structure. First, we apply three-fold cross-fitting of a standard machine learning model so we have a ground-truth value of $Y$ and a prediction $\widehat{Y}$ for each point. Then we add the adversarial constraint and conduct the same cross-fitting procedure. Finally, we estimate our regression of interest using the ground truth data, the baseline predictions, and adversarial model predictions. We test both the simple linear regression adversary described above (SLR adversary) as well as a model that penalizes the absolute value of $Cov(X, \nu)$ (the correlational adversary). We also test our bias correction method using the baseline model predictions. For all model-based predictions we bootstrap standard errors to include the uncertainty from training the model.

In this simulation and the next simple example, we focus on the cross-sectional relationship between roads and forest cover. Previous work has found that roads are an important driver of deforestation \citep{asher_ecological_2020}. Roads and other infrastructure are non-randomly placed, so this cross-sectional relationship is likely to generate confounder-induced bias (Supplemental Figure \ref{fig:examples}.c). Consider $X$ to be construction of a road, $Y$ to be forest cover, and $W$ to be some omitted geographic variable, like slope, that influences both measurement errors and the placement of roads.

Our first simulation follows this procedure:
\begin{enumerate}
    \item Draw 20,000 observations of $W$ from a Poisson distribution with shape parameter of 1. 
    \item Assign each observation $X \sim$ Bernoulli with $p = \max(1-W/4, 0)$, so that treatment is more likely when $W$ is lower.
    \item Assign each observation a random forest cover $Y$ and associated satellite data $k$ from the Bastin points. 
    \item If $W>0$, make the satellite data artificially `greener' without changing the true forest cover label. In practice this is done by replacing $k$ with the satellite data from a different point with a higher percent forest cover.
\end{enumerate}

The true treatment effect of $X$ is zero, since the forest labels $Y$ are assigned randomly. However the last step mimics a real source of bias --- remotely sensed forest cover tends to be overestimated on steeper slopes since images are taken from above and tend to capture more trees in a smaller spatial area when on a slope. Because of this bias, and selection into treatment, it will appear that roads are associated with lower forest cover. Note also that there are no ``traditional'' confounders here --- nothing in the simulation is associated with both road proximity and true forest cover.

\begin{figure}[h!]
    \centering
    \includegraphics[width = \linewidth]{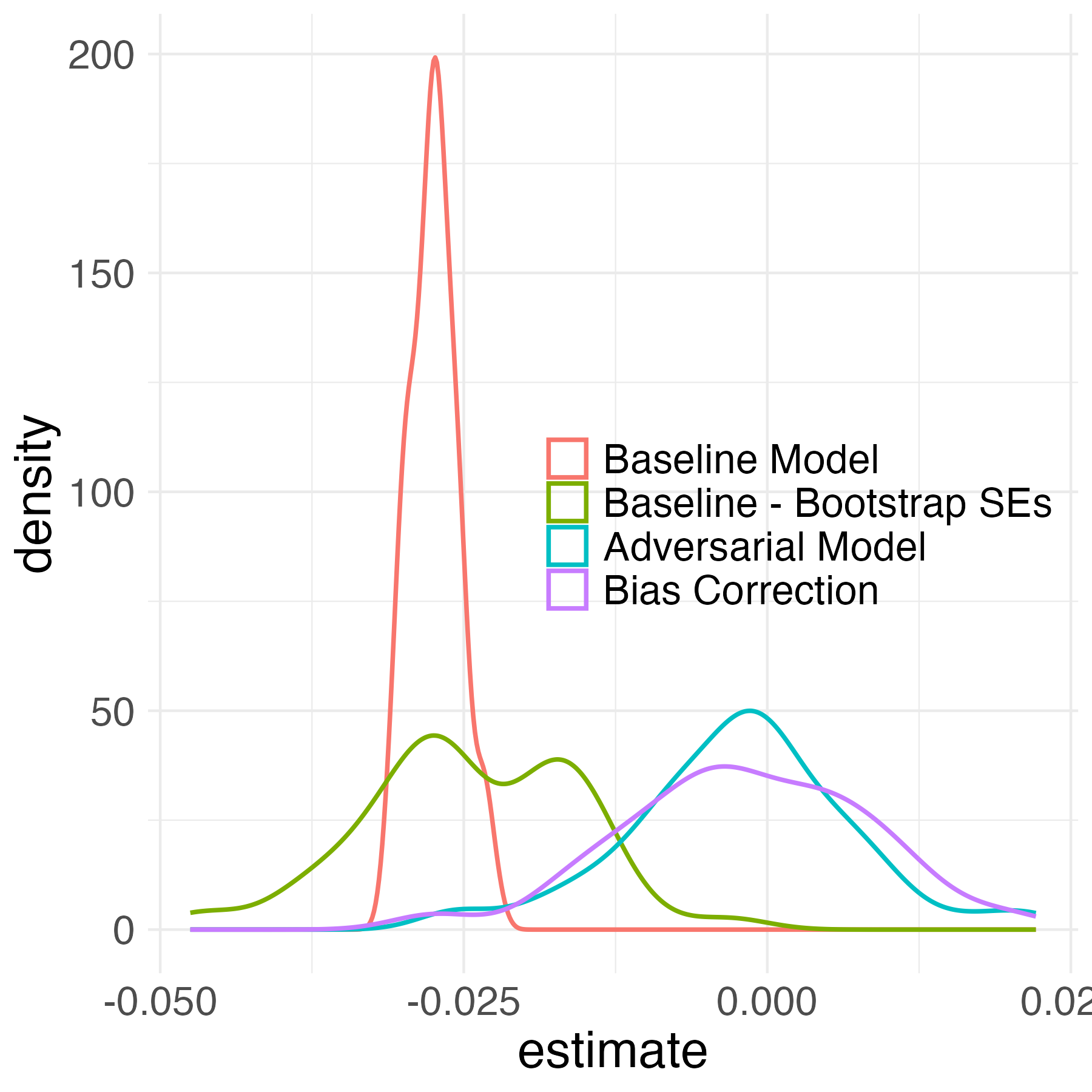}
    \caption{Estimates from the baseline model vs the adversarial models with 10,000 labeled observations. Each distribution represents the distribution of the coefficients from each model after 100 runs on bootstrapped training data.}
    \label{fig:progressive_end}
\end{figure}

Figure \ref{fig:progressive_end} shows the distribution of coefficient estimates from 100 bootstrapped runs of each of the models and regressions using 10,000 training points. As predicted, using the baseline machine learning model results in a negative and significant estimate of the effect of $X$ on $Y$. Failing to bootstrap standard errors that include prediction model uncertainty (e.g. using satellite data products as proxies for true values) leads to dramatically overstated precision. Both the bias correction method and the adversarial model result in coefficient distributions correctly centered around 0.  

We test the performance of each model using a progressively increasing sample of labeled points. For each given sample size of labeled points $J$, we use the $J$ labels to train the model, and then predict on the remaining points so that the sample size for the regression is always $N = 20,000$. For each $J$ we bootstrap 100 different versions of the model to estimate standard errors as well. 

\begin{figure}[h!]
    \centering
    \includegraphics[width = \linewidth]{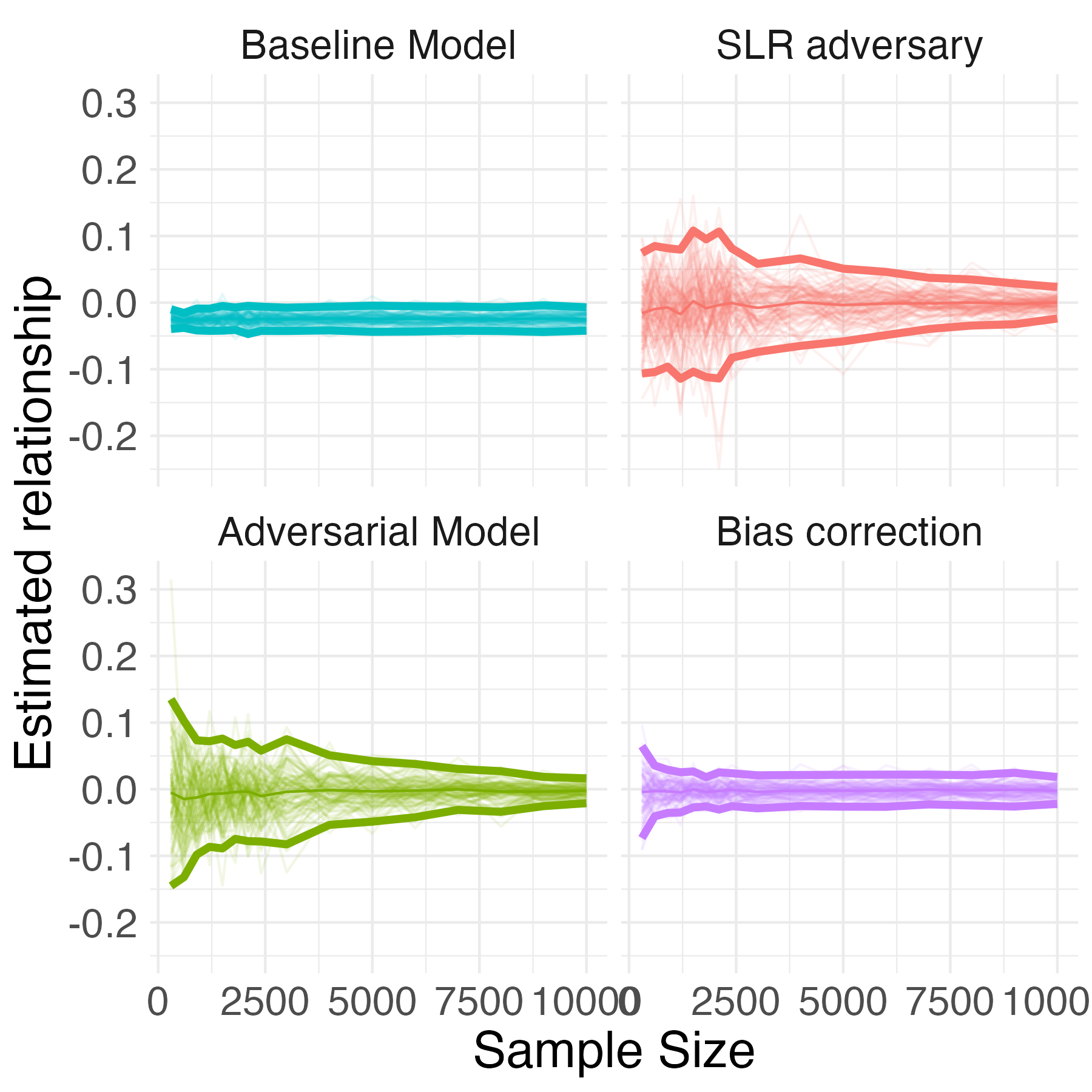}
    \caption{Estimates from the baseline model vs the adversarial models across sample sizes. Each light colored line is an individual training run where researchers label progressively more observations. The thick lines represent the mean and two standard deviations from the mean of the runs.}
    \label{fig:progressive}
\end{figure}

Figure \ref{fig:progressive} shows the results of this exercise. The baseline model learns to measure forest cover with relatively few training observations but generates biased estimates. The adversarial models have much higher variances at small sample sizes but are consistently centered on $\beta = 0$, and precision increases as the sample size of labeled points increases. This bias correction method works well at smaller sample sizes and consistently recovers the true null effect. 

In this setting the bias correction method performs best across smaller sample sizes because the satellite data contains no information about the source of the bias, since it has been replaced by imagery from randomly drawn pixels with higher forest cover. This means that the adversarial model has to do a worse job predicting the low $W$, high $Y$ observations so that the measurement error is balanced across $X$. Note that the adversary never has access to $W$, yet is still able to adjust for $W$-induced measurement error. In real world cases we expect the adversarial model to outperform the bias correction method by learning representations that predict bias and adjusting for them.

\begin{figure}[h!]
    \centering
    \includegraphics[width = \linewidth]{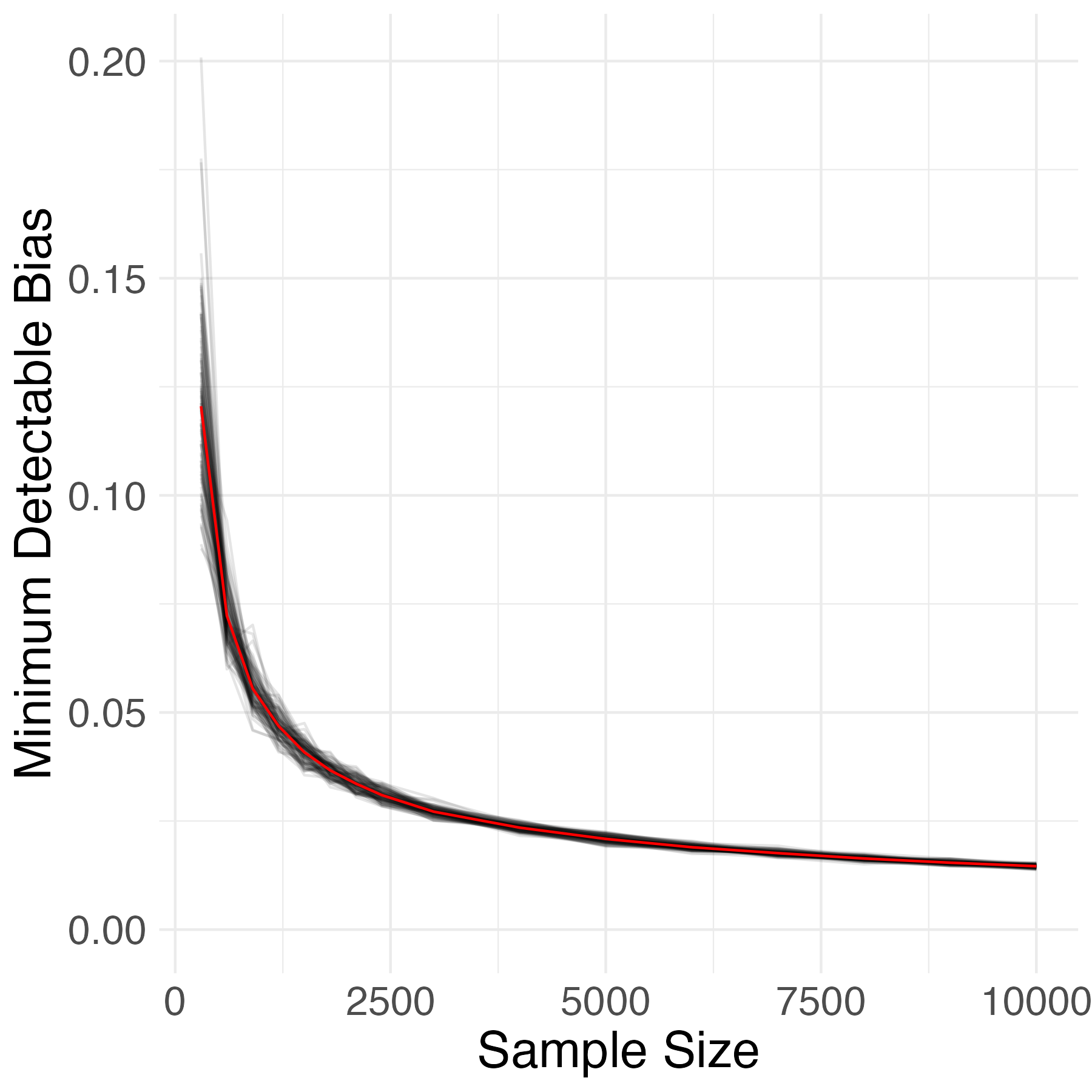}
    \caption{Minimum detectable bias (MDB) across sample sizes at power of 0.8 and $\alpha = 0.05$. Each black line represents an estimate of MDB using a different random samples of labeled data, red line is the true MDB using standard errors estimated with the whole dataset.}
    \label{fig:power}
\end{figure}

Finally, we conduct power analyses of our bias test to estimate the minimum detectable bias (MDB) at different sample sizes. The results are presented in Figure \ref{fig:power}. Each line represents a different random draw of points to label. At each sample size, for each set of points, we estimate the standard error of $\gamma$ and use that to perform a standard power calculation using 0.8 power and 95\% statistical significance. Given that the true magnitude of the bias is 0.025 in this simulation, researchers would need to label more than 2,500 points to detect this bias as statistically different from zero 80\% of the time. For researchers whose standard impact evaluation procedures result in small effect sizes, it is relevant to understand the smallest bias that could be detected by the bias test, because failing to reject that the bias is not zero with a low-powered test is not very reassuring.

\subsection{Descriptive Exercise: Roads and Forest Cover} \label{sec:xsection}

\begin{figure}[h!]
    \centering
    \includegraphics[width = \linewidth]{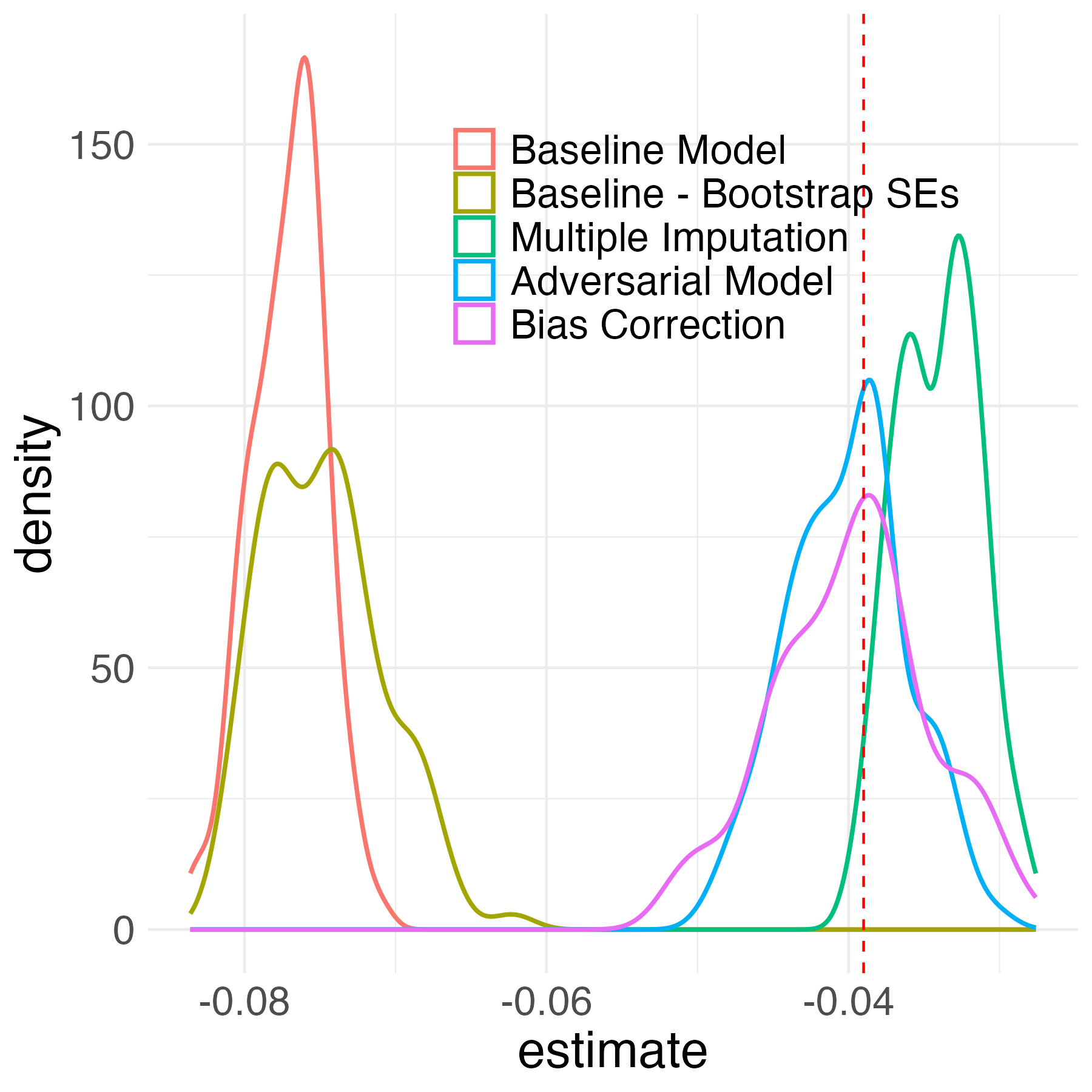}
    \caption{Estimates from the baseline model vs the adversarial models with 10,000 labeled observations. Each distribution represents the distribution of the coefficients from each model after 100 runs on bootstrapped training data. The dashed red vertical line represents estimate using all ground truth labels.}
    \label{fig:roadsdists}
\end{figure}

Next we use the true \cite{bastin_extent_2017} data, and data on the African road network from \citet{meijer_global_2018}, to estimate the gradient of forest cover with respect to distance to the nearest road. Whereas before the independent variable was binary and the outcome was continuous, now our independent variable is continuous, log distance to the nearest road, and our outcome is binary (forested or not). We apply the same cross-fitting procedure as above for a standard model, an adversarial model using a 3-layer neural network, the bias correction approach, and the multiple imputation method recommended by \citet{proctor_parameter_2023}. We then run the same regressions as in Section \ref{sec:sim}. The results are shown in Figure \ref{fig:roadsdists}. 

The standard machine-learning model overestimates the negative relationship between proximity to roads and forest cover. However, the adversarial model and the bias correction method are both able to generate measurements that recover the true estimate. The multiple imputation approach is precise, but still somewhat biased. In this setting the adversarial model produces more precise estimates than the bias correction approach. 

\begin{figure}[h!]
    \centering
    \includegraphics[width = .8\linewidth]{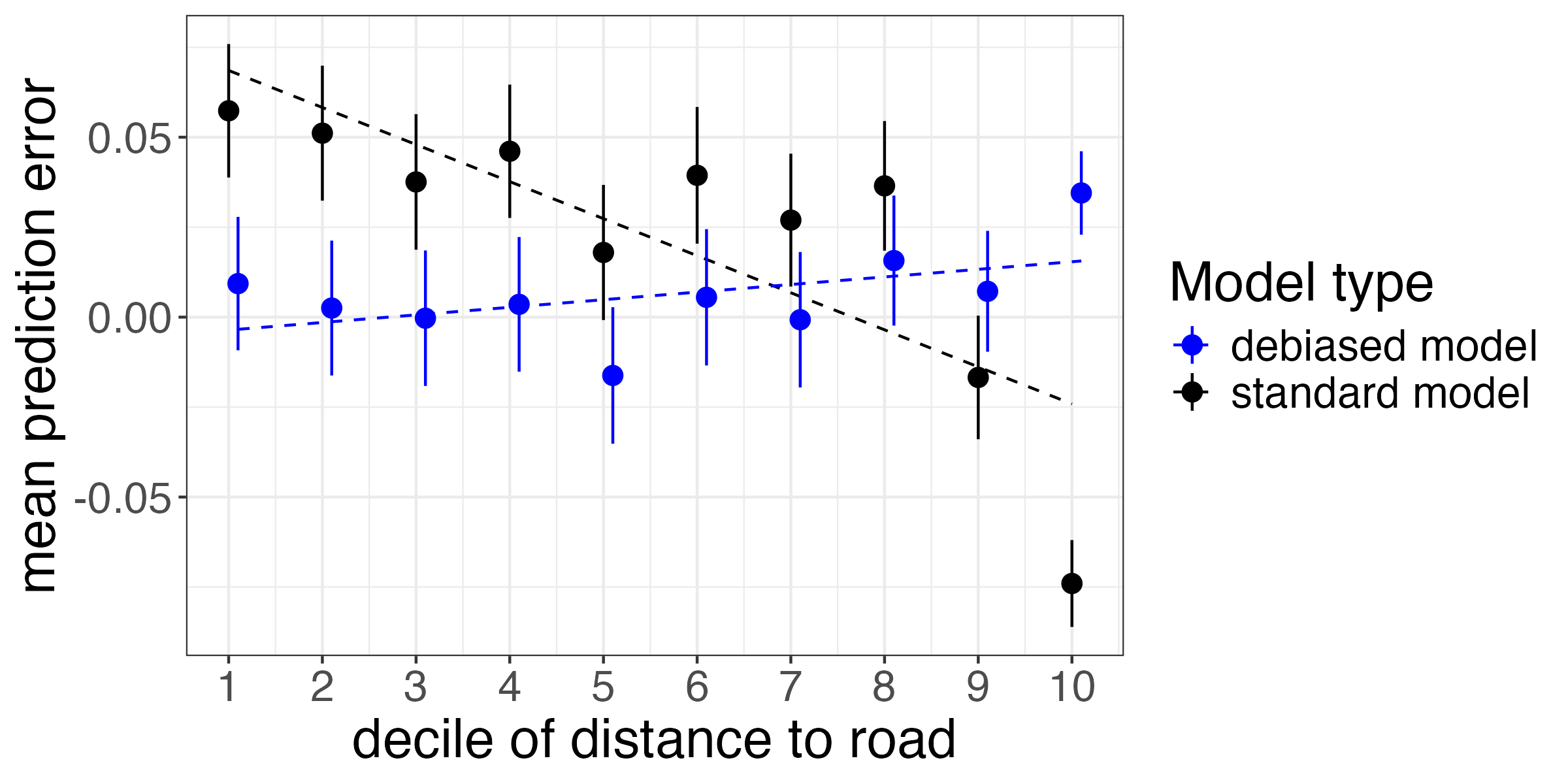}
    \caption{Measurement error across deciles of distance to road}
    \label{fig:resids}
\end{figure}

We investigate the source of the bias in Figure \ref{fig:resids}, where we plot the mean measurement error at each decile of distance from a road for both the adversarial model and the standard model. Both models have an average of 0 prediction error (by construction), but the standard model produces errors with a gradient with respect to the treatment variable while the adversarial model does not.

\begin{figure*}[h]
    \centering
    \begin{subfigure}{0.45\textwidth}
        \includegraphics[width=\textwidth]{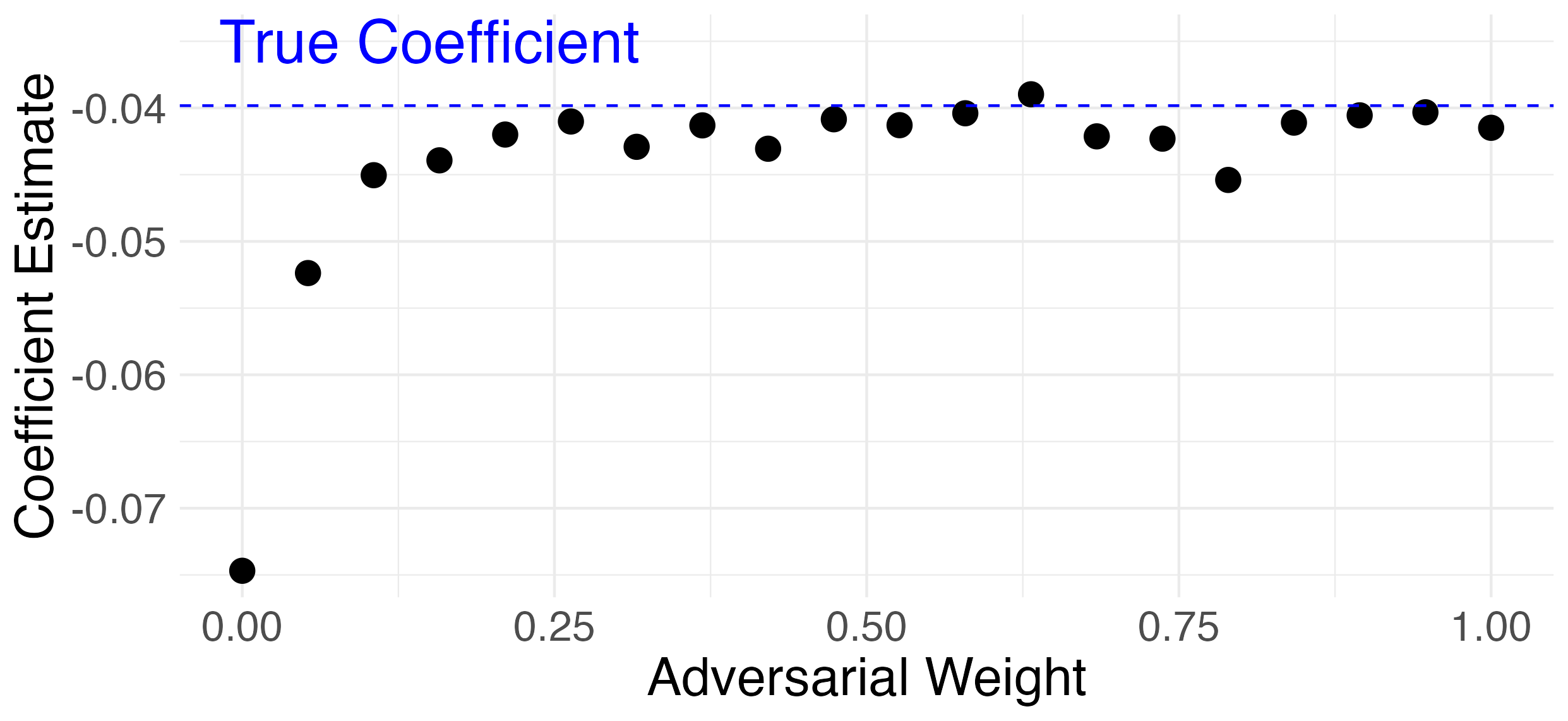}
        \caption{Bias: Logistic Regression Primary Model}
    \end{subfigure}
    \hfill
    \begin{subfigure}{0.45\textwidth}
        \includegraphics[width=\textwidth]{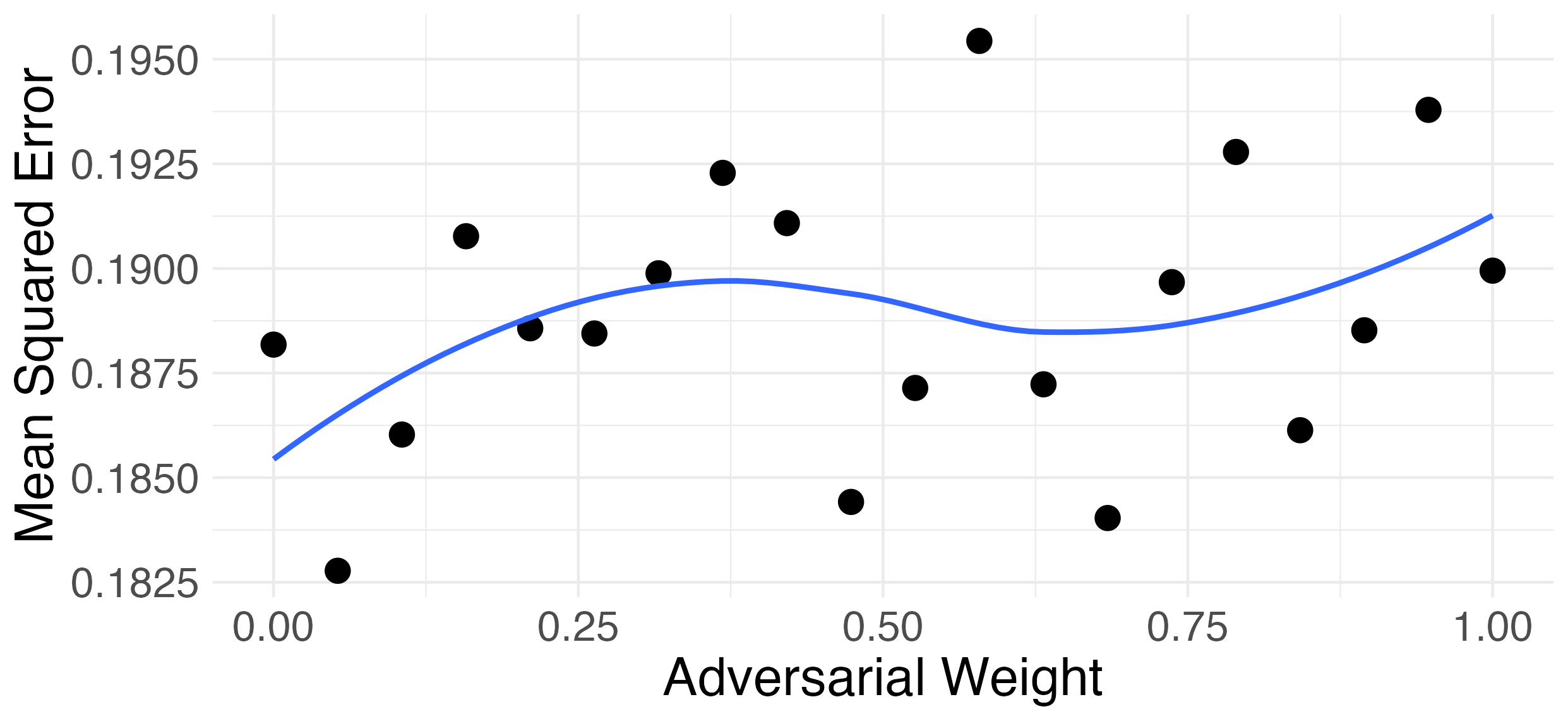}
        \caption{Accuracy: Logistic Regression Primary Model}
    \end{subfigure} \\
    \begin{subfigure}{0.45\textwidth}
        \includegraphics[width=\textwidth]{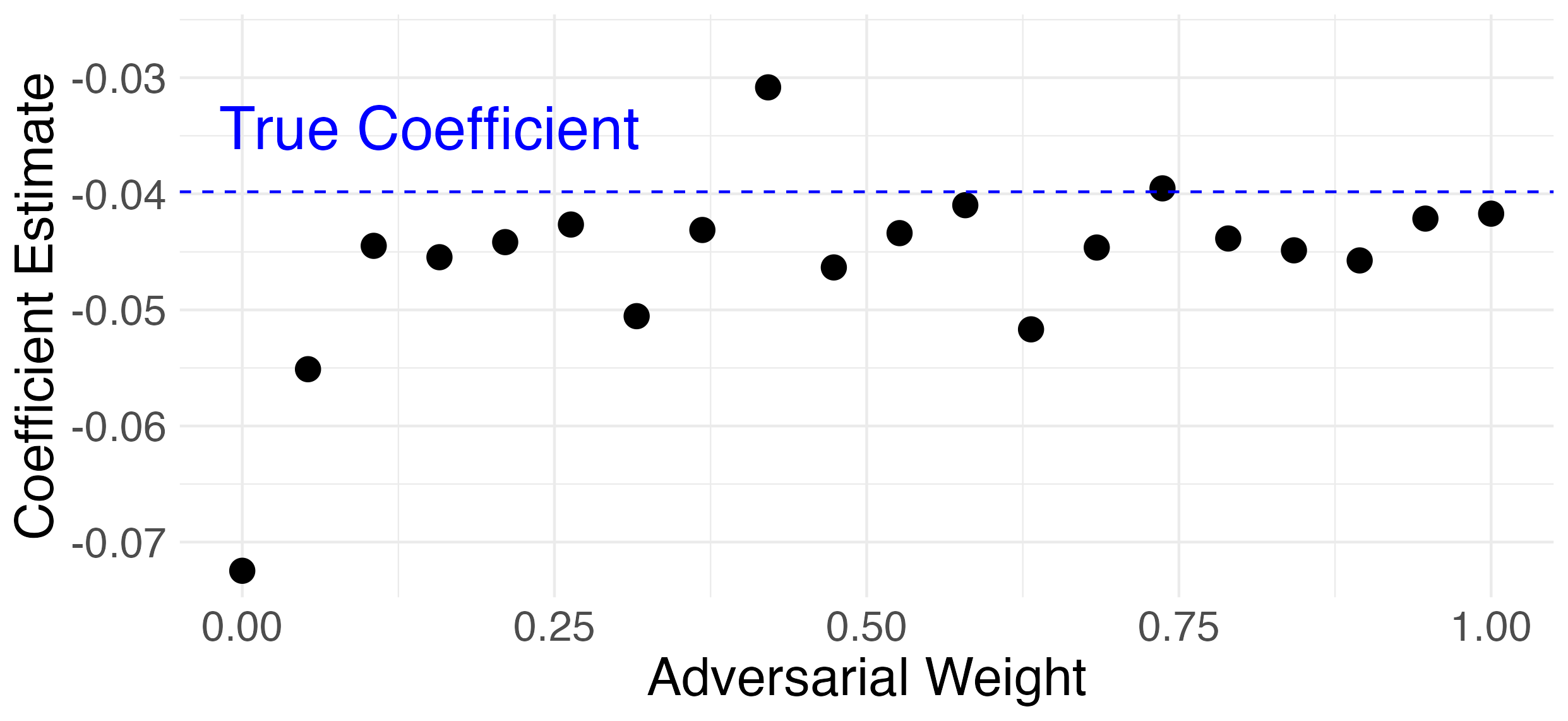}
        \caption{Bias: DNN Primary Model}
    \end{subfigure}
    \hfill
    \begin{subfigure}{0.45\textwidth}
        \includegraphics[width=\textwidth]{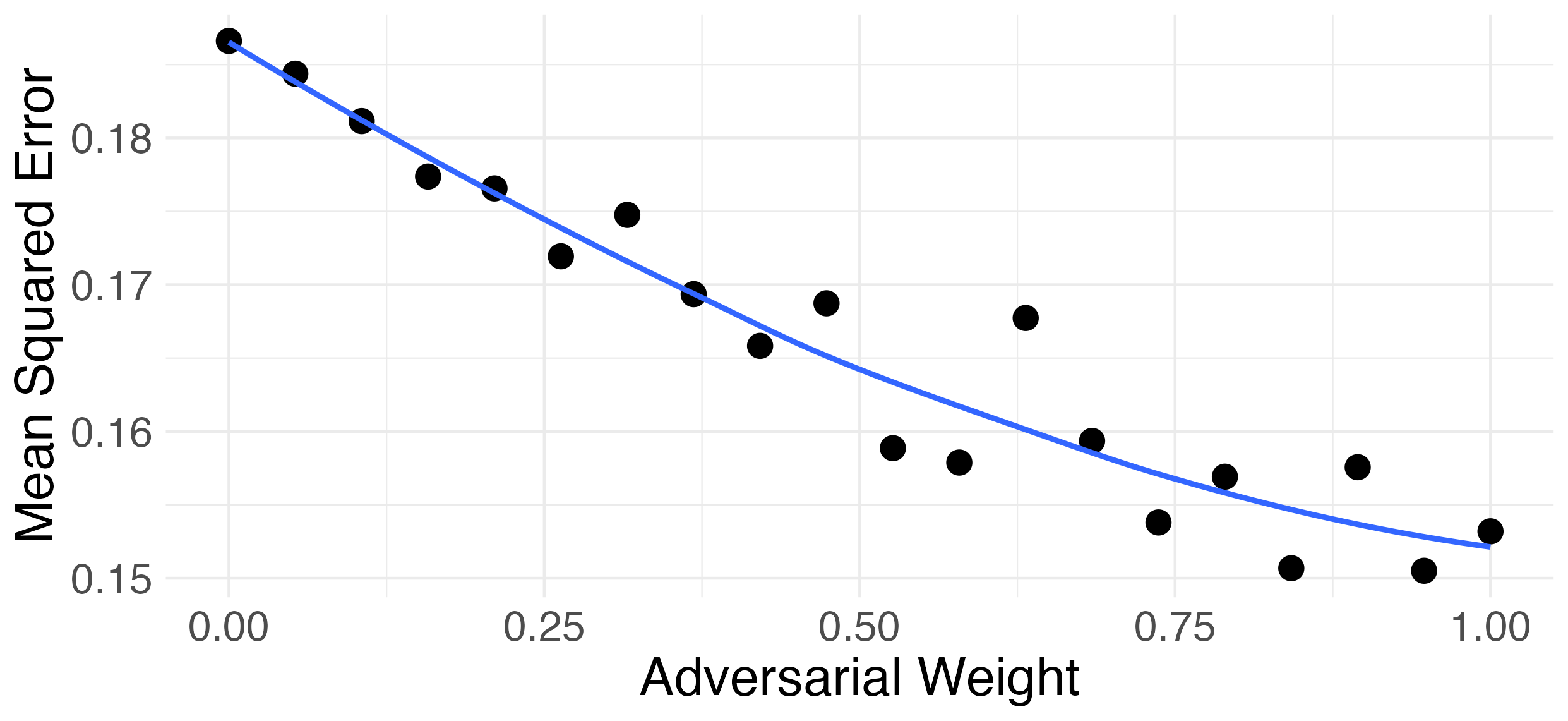}
        \caption{Accuracy: DNN Primary Model}
    \end{subfigure}
    \caption{Tuning $\alpha$: Tradeoffs between bias and accuracy. Graphs show how coefficient bias (left column) and overall predictive accuracy (MSE - right column) change as the weight on the adversary increases. Top row shows results for a primary model that is a logistic regression. Bottom row shows results for a Deep Neural Net (DNN - 3 layers). Blue lines show Loess smoothed best fit curves.}
    \label{fig:tuning}
\end{figure*}

We use this setting to run experiments on the $\alpha$ parameter: the researcher chosen weight on the adversary in the loss function. The results are summarized in Figure \ref{fig:tuning} for two different primary prediction models, a logistic regression, and a deep neural net (DNN). The left column shows that increasing the weight on the adversary from zero (standard model) to 1 quickly eliminates bias in the estimated coefficients. We expected the MSE to increase at higher $\alpha$ values, and this is the case for the logistic primary model. For the DNN, however, increasing the adversary's weight improves prediction accuracy. This result can be understood as a kind of regularization effect. In some settings with many model parameters, it is well known that a regularization penalty can reduce overfitting and improve out-of-sample prediction performance \citep{tibshirani_regression_1996}. While it is difficult to say precisely when adversarial models will improve prediction accuracy, the adversary seems to be serving a regularization function in this example.

\section{Discussion}
\label{discussion} 
We have introduced an adversarial machine learning approach to debiasing predictions for causal inference. Our results show that it effectively debiases predictions in a simulation where the bias is artificially constructed, and in an application where researchers estimate the relationship between roads and forest cover. We also construct a test for measurement error bias and a guidelines for researchers to estimate the number of observations they will need to label to detect and remove the bias. We also show that this approach can succeed where competing approaches fail. This debiasing algorithm can be applied to any setting where machine-learned predictions will be used as an outcome variable for a causal task and does not require the researcher to have any knowledge of what may be causing measurement error bias.

There are two main limitations to our approach. First, it requires researchers to build an adversarial model which to our knowledge is only feasible for methods that use stochastic gradient descent-based optimization algorithms. Within this set of machine learners, the addition of the adversarial component can be challenging to implement in common machine learning libraries. Second, a single run of the algorithm does not always achieve good results. In experiments we found that causal parameters from the adversarial model are closer to the true parameter (compared to a standard model) most of the time, sometimes researchers can achieve the best performance by only using hand-labeled observations or by using a standard model. We suspect this instability is due to a lack of smoothness in the gradient of the loss function when the adversary is added.

In future work we plan to address the second issue to see when adjustments to the optimization algorithm or adversarial functional form can achieve more stable results. As researchers in remote sensing or other computational fields increasingly turn to constructing their own measurement models, we expect our approach to become more useful compared to approaches that debias existing predictions.




\section*{Acknowledgements}
Thanks to Eli Fenichel, Philipp Ketz, the Berkeley Political Methodology Seminar, The Washington University St Louis Political Methodology Seminar, the LSE/Imperial College Workshop, and participants at the AGU, ICLR, TWEEDS, PolMeth, and AERE conferences for helpful feedback and support.

\bibliographystyle{unsrtnat}
\bibliography{references.bib}

\newpage
\appendix
\onecolumn
\section{Proof that maximizing adversarial loss decreases bias when accuracy is held constant} \label{sec:appendixa}

Define $P = X(X'X)^{-1}X'$ as the n x n symmetric and idempotent projection matrix, and $\mathbb{I}$ as the n x n identity matrix. The adversary's loss function is:

\begin{gather}
    L_a = (\nu - P\nu)'(\nu - P\nu) \\ \nonumber
     = v'(\mathbb{I} - P)'(\mathbb{I} - P)\nu \\ \nonumber
     = v'(\mathbb{I} - P)\nu = \nu'\nu - \nu'P\nu 
\end{gather}

by the properties of the projection matrix. Consider prediction errors at step $t$ and $t + 1$ of training: $\nu_t$ and $\nu_{t + 1}$, and suppose that their prediction accuracy is equivalent. Maximizing the adversarial loss implies that $L_a(\nu_{t + 1}) > L_a(\nu_t)$. Therefore,

\begin{gather}
    \nu_{t + 1}'\nu_{t + 1} - \nu_{t + 1}'P\nu_{t + 1} > \nu_t'\nu_t - \nu_t'P\nu_t \\ \nonumber
    \nu_{t + 1}'P\nu_{t + 1} < \nu_t'P\nu_t \nonumber \\
    \nu_{t + 1}'PP\nu_{t + 1} < \nu_t'PP\nu_t \\
    \nu_{t + 1}'X(X'X)^{-1}X'X(X'X)^{-1}X'\nu_{t + 1} < \nu_t'X(X'X)^{-1}X'X(X'X)^{-1}X'\nu_t \nonumber \\
    \gamma_{t + 1} X'X \gamma_{t + 1} < \gamma_t X'X \gamma_t \nonumber \\
    |\gamma_{t + 1}| < |\gamma_t|.
\end{gather}






Concluding the proof.

\clearpage

\section{Debiasing with Control Variables and Instruments} \label{sec:appendixb}

\subsection*{Adding Control Variables}

Assume we want to estimate $\beta_1$ in the regression

\begin{equation}
    \widehat{Y}_i = \beta_1 x_1 + x_2 \beta_2 + e_i
\end{equation}

where $x_1$ is an n x 1 vector of the treatment variable, and $x_2$ is an n x k matrix of control variables. By the Frisch-Waugh-Lovell theorem, we can write $\widehat{\beta}_1$ as:

\begin{equation}
    \widehat{\beta}_1 = (\Tilde{X}_1 ' \Tilde{X}_1)^{-1} \Tilde{X}_1 ' \Tilde{\widehat{Y}}
\end{equation}

\noindent where $\Tilde{X}_1$ are the residuals of the regression of $X_1$ on $X_2$, and $\Tilde{\widehat{Y}}$ are the residuals of the regression of $\widehat{Y}$ on $X_2$. $\widehat{\beta}_1$ can thus be rewritten as follows:

\begin{gather}
    \widehat{\beta}_1 = (\Tilde{X}_1 ' \Tilde{X}_1)^{-1} \Tilde{X}_1 ' (\mathbb{I} - X_2 (X_2 ' X_2)^{-1} X_2') (Y + \nu) \\  \nonumber 
    = (\Tilde{X}_1 ' \Tilde{X}_1)^{-1} \Tilde{X}_1 ' (\mathbb{I} - X_2 (X_2 ' X_2)^{-1} X_2') Y + \\  \nonumber
   (\Tilde{X}_1 ' \Tilde{X}_1)^{-1} \Tilde{X}_1 ' (\mathbb{I} - X_2 (X_2 ' X_2)^{-1} X_2') \nu \\  \nonumber
   = (\Tilde{X}_1 ' \Tilde{X}_1)^{-1} \Tilde{X}_1 ' \Tilde{Y} + (\Tilde{X}_1 ' \Tilde{X}_1)^{-1} \Tilde{X}_1 ' \Tilde{\nu}  \nonumber
\end{gather}

\noindent where $\Tilde{\nu}$ are the residuals of the regression of $\nu$ on $X_2$. The expectation of this estimate is

\begin{equation}
    \mathbb{E}[\beta_1] = \beta_1 + \frac{cov(\Tilde{X}_1, \Tilde{\nu})}{var(\Tilde{X}_1)}.
\end{equation}

Intuitively this makes sense -- if the residual variation in $X_1$ is correlated with the residual prediction error, after controlling for $X_2$ in both cases, our estimate will be biased. Thus following the same logic as above, we can make the adversary a linear regression of $\Tilde{\nu}$ on $\Tilde{X}_1$. The same logic applies for when adding additional covariates to the regression. 

\subsection*{Instrumental Variables}

A similar argument can be extended to the instrumental variables case with controls. Following the two-stage least squares estimation procedure, we first use covariates $X_2$ and instruments $Z$ to predict $X_1$:

\begin{equation}
    \widehat{X}_1^{IV} = C(C'C)^{-1}C'X_1
\end{equation}

where $C = [1, X_2, Z]$. Then, we regress the outcome against the predicted values of $X_1$ and the covariates $X_2$ and take the estimated coefficient for $\widehat{X}_1^{IV}$ in this second stage regression as our estimate of the true $\beta_1$. By the Frisch-Waugh-Lovell theorem, we have

\begin{align}
    \widehat{\beta_1}^{2SLS} &= (\Tilde{\widehat{X}_1}'\Tilde{\widehat{X}_1})^{-1} \Tilde{\widehat{X}_1}'\Tilde{\widehat{Y}} \\ \nonumber
    &= (\Tilde{\widehat{X}_1}'\Tilde{\widehat{X}_1})^{-1} \Tilde{\widehat{X}_1}'\Tilde{Y} + (\Tilde{\widehat{X}_1}'\Tilde{\widehat{X}_1})^{-1} \Tilde{\widehat{X}_1}'\Tilde{\nu}
\end{align}

where $\Tilde{\widehat{X_1}}$ are the residuals from regressing $\widehat{X}_1^{IV}$ on $X_2$, $\Tilde{Y}$ are the residuals from regressing $Y$ on $X_2$, and $\Tilde{\nu}$ are the residuals from regressing $\nu$ on $X_2$.

In the case of a single instrument $Z$, this estimator of $\beta_1$ can be rewritten more simply following the indirect least-squares procedure. For this approach, we perform linear regressions for the models 

\begin{equation}
    \widehat{Y}_i = \gamma_0 + \gamma_1 w_i + \gamma_2' X_{2i} + w_i;
\end{equation}
\begin{equation}
    \widehat{X}_{1i} = \alpha_0 + \alpha_1 Z_i + \alpha_2' X_{2i} + u_i
\end{equation}

This produces the following estimate of $\beta_1$, which coincides with $\widehat{\beta_1}^{2SLS}$ for this special case: 

\begin{align}
    \widehat{\beta}_1^{ILS} &= \frac{\widehat{\gamma_1}}{\widehat{\alpha_1}} 
    = \frac{(\Tilde{Z}'\Tilde{Z})^{-1} \Tilde{Z}'\Tilde{\widehat{Y}}}{(\Tilde{Z}'\Tilde{Z})^{-1} \Tilde{Z}'\Tilde{X_1}} 
    = \frac{cov(\Tilde{Z}, \Tilde{\widehat{Y}})}{cov(\Tilde{Z}, \Tilde{X_1})} \\ \nonumber
    &= \frac{cov(\Tilde{Z}, \Tilde{Y})}{cov(\Tilde{Z}, \Tilde{X_1})} + \frac{cov(\Tilde{Z}, \Tilde{\nu})}{cov(\Tilde{Z}, \Tilde{X_1})}
\end{align}

In the above expressions for $\widehat{\beta_1}^{2SLS}$ and $\widehat{\beta}_1^{ILS}$, we see that the coefficient estimates are the sum of the coefficient estimate that we would obtain from performing these procedures given $Y$ without measurement error - which is consistent for $\beta_1$ given IV assumptions - and an additional bias term involving the measurement error $\nu$. To minimize this bias, we propose an adversary in the form of a regression of $\Tilde{\nu}$ on $\Tilde{Z}$ for the single instrument case, or $\Tilde{\nu}$ on $\Tilde{\hat{X_1}}$ more generally.

\section{Adversarial Debiaser Algorithm} \label{sec:appendixc}

\begin{algorithm}[ht!]
\caption{Adversarial Debiasing Algorithm}
\label{alg:adversarial_debiasing}
\begin{algorithmic}[1]

  \Require Labeled data $\{(k_j,x_j,y_j)\}_{j=1}^N$; 
           primary model $P(\cdot;\omega)$; 
           adversary $A(\cdot;\gamma)$; 
           loss functions $L_p$ and $L_a$; 
           learning rates $\eta_\omega,\ \eta_\gamma$; 
           adversarial weight $\alpha$; 
           number of training iterations $T$.

  \State Initialize primary model parameters $\omega$.
  \State Initialize adversary parameters $\gamma$.

  \For{$t = 1,\dots,T$}
    \State \textbf{Primary Model Forward Pass:}
    \State For each sample $j$, compute $\widehat{Y}_j \leftarrow P(k_j; \omega)$.
    \State Compute prediction errors $\nu_j \leftarrow y_j - \widehat{Y}_j(\omega)$.

    \State \textbf{Adversary Update (minimize $L_a$):}
    \State $\gamma \leftarrow \gamma - \eta_\gamma \nabla_{\gamma} \bigl[ L_a(x_j,\nu_j,\gamma) \bigr]$.

    \State \textbf{Primary Model Update (minimize $L_p - \alpha\,L_a$):}
    \State $\omega \leftarrow \omega - \eta_\omega \nabla_{\omega} 
            \Bigl[L_p(\widehat{Y},\,y) - \alpha\,L_a(x_j,\nu_j,\gamma)\Bigr]$.
  \EndFor

  \Return Trained primary model parameters $\omega^*$

\end{algorithmic}
\end{algorithm}

\clearpage

\section{Additional Figures}

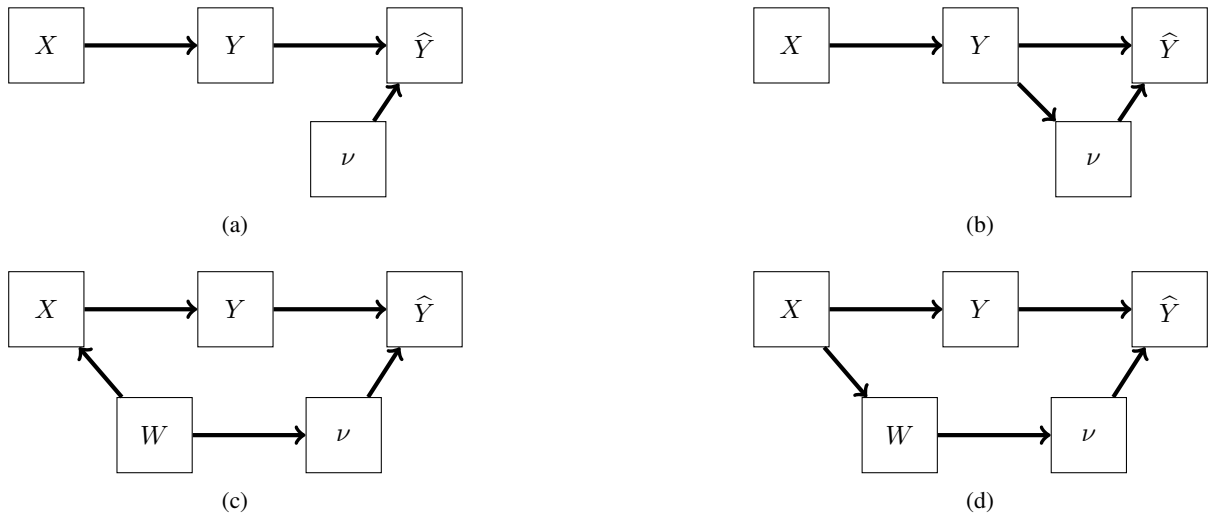
\begin{figure*}[h]
  \centering
  \begin{subfigure}{0.4\linewidth}
    \centering
    \begin{tikzpicture}[node distance=1.5cm, minimum size=1cm, font=\bfseries]
        \node[draw, rectangle] (X) {$X$};
        \node[draw, rectangle, right=of X] (Y) {$Y$};
        \node[draw, rectangle, right=of Y] (Yhat) {$\widehat{Y}$};
        \node[draw, rectangle, below=of Y, xshift=1.5cm, yshift=1cm] (nu) {$\nu$};
    
        \draw[->, ultra thick] (X) -- (Y);
        \draw[->, ultra thick] (Y) -- (Yhat);
        \draw[->, ultra thick] (nu) -- (Yhat);
    \end{tikzpicture}
    \caption{}
  \end{subfigure}
  \hfill
  \begin{subfigure}{0.4\linewidth}
    \centering
    \begin{tikzpicture}[node distance=1.5cm, minimum size=1cm, font=\bfseries]
        \node[draw, rectangle] (X) {$X$};
        \node[draw, rectangle, right=of X] (Y) {$Y$};
        \node[draw, rectangle, right=of Y] (Yhat) {$\widehat{Y}$};
        \node[draw, rectangle, below=of Y, xshift=1.5cm, yshift=1cm] (nu) {$\nu$};
    
        \draw[->, ultra thick] (X) -- (Y);
        \draw[->, ultra thick] (Y) -- (nu);
        \draw[->, ultra thick] (Y) -- (Yhat);
        \draw[->, ultra thick] (nu) -- (Yhat);
    \end{tikzpicture}
    \caption{}
  \end{subfigure}
  \vskip\baselineskip
    \begin{subfigure}{0.4\linewidth}
    \centering
    \begin{tikzpicture}[node distance=1.5cm, minimum size=1cm, font=\bfseries]
        \node[draw, rectangle] (X) {$X$};
        \node[draw, rectangle, right=of X] (Y) {$Y$};
        \node[draw, rectangle, right=of Y] (Yhat) {$\widehat{Y}$};
        \node[draw, rectangle, below left=of Y, xshift=1cm, yshift=0.4cm] (W) {$W$};
        \node[draw, rectangle, below left=of Yhat, xshift=1cm, yshift=0.4cm] (nu) {$\nu$};
    
        \draw[->, ultra thick] (X) -- (Y);
        \draw[->, ultra thick] (Y) -- (Yhat);
        \draw[->, ultra thick] (nu) -- (Yhat);
        \draw[->, ultra thick] (W) -- (X);
        \draw[->, ultra thick] (W) -- (nu);
    \end{tikzpicture}
    \caption{}
  \end{subfigure}
  \hfill
  \begin{subfigure}{0.4\linewidth}
    \centering
    \begin{tikzpicture}[node distance=1.5cm, minimum size=1cm, font=\bfseries]
        \node[draw, rectangle] (X) {$X$};
        \node[draw, rectangle, right=of X] (Y) {$Y$};
        \node[draw, rectangle, right=of Y] (Yhat) {$\widehat{Y}$};
        \node[draw, rectangle, below left=of Y, xshift=1cm, yshift=0.4cm] (W) {$W$};
        \node[draw, rectangle, below left=of Yhat, xshift=1cm, yshift=0.4cm] (nu) {$\nu$};
    
        \draw[->, ultra thick] (X) -- (Y);
        \draw[->, ultra thick] (Y) -- (Yhat);
        \draw[->, ultra thick] (nu) -- (Yhat);
        \draw[->, ultra thick] (X) -- (W);
        \draw[->, ultra thick] (W) -- (nu);
    \end{tikzpicture}
    \caption{}
  \end{subfigure}
  \caption{Four Directed-Acrylic-Graphs illustrating potential relationships between treatment ($X$), outcomes ($Y$), measurement error ($\nu$), machine learning model predictions ($\widehat{Y}$), and unobserved variables ($W$). (a) is classical measurement error, (b) shows outcome-induced bias, (c) shows confounder-induced bias, and (d) shows treatment-induced bias.}
    \label{fig:examples}
\end{figure*}

\end{document}